\documentclass[11pt,a4paper]{article}
\usepackage[hyperref]{emnlp-ijcnlp-2019}
\usepackage{times}
\usepackage{latexsym}

\usepackage{url}

\usepackage{algorithmicx}  
\usepackage{algpseudocode}  
\usepackage{amsmath}  
\usepackage{color,xcolor}
\usepackage{algorithm} %format of the algorithm 

\usepackage{pifont}
\usepackage{bm}
\usepackage{amsfonts}

\usepackage{graphicx}
\usepackage[normalem]{ulem}
\usepackage{multirow}
\usepackage{amsmath}
\usepackage{url}

\usepackage{amsmath}
\usepackage{amssymb}
\usepackage{amsopn}
\usepackage{subcaption}
\usepackage{graphicx}
\usepackage{multirow}
\usepackage[english]{babel}
\usepackage{bm}
\usepackage{array}

\aclfinalcopy % Uncomment this line for the final submission

\title{Iterative Dual Domain Adaptation for Neural Machine Translation}

\author{Jiali Zeng$^{1}$, \ Yang Liu$^{2}$, \  Jinsong Su$^{1}\thanks{\ \ Corresponding author.}$, \
Yubin Ge$^{3}$, \ Yaojie Lu$^{4}$, \ Yongjing Yin$^{1}$, \ Jiebo Luo$^{5}$\\
$^{1}$Xiamen University, Xiamen, China \ \ \ $^{2}$Tsinghua University, Beijing, China\\
$^{3}$University of Illinois at Urbana-Champaign, Urbana, IL 61801, USA\\
$^{4}$Institute of Software, Chinese Academy of Sciences, Beijing, China\\
$^{5}$Department of Computer Science, University of Rochester, Rochester NY 14627, USA\\
 {\tt lemon@stu.xmu.edu.cn  liuyang2011@tsinghua.edu.cn} \\
 {\tt jssu@xmu.edu.cn } \\ 
}

\date{}

\begin{document}
\maketitle
\begin{abstract}
		
Previous studies on the domain adaptation for neural machine translation (NMT) mainly focus on the one-pass transferring out-of-domain translation knowledge to in-domain NMT model.
In this paper, 
we argue that such a strategy fails to fully extract the domain-shared translation knowledge, 
and repeatedly utilizing corpora of different domains can lead to better distillation of domain-shared translation knowledge.
To this end, 
we propose an iterative dual domain adaptation framework for NMT.
Specifically,
we first pre-train in-domain and out-of-domain NMT models using their own training corpora respectively, 
and then iteratively perform bidirectional translation knowledge transfer (from in-domain to out-of-domain and then vice versa) based on knowledge distillation until the in-domain NMT model convergences.
Furthermore, 
we extend the proposed framework to the scenario of multiple out-of-domain training corpora,
where the above-mentioned transfer is performed sequentially between the in-domain and each out-of-domain NMT models in the ascending order of their domain similarities.
Empirical results on Chinese-English and English-German translation tasks demonstrate the effectiveness of our framework.
\end{abstract}
	
\section{Introduction}
Currently, 
neural machine translation (NMT) has become dominant in the community of machine translation due to its excellent performance \cite{Bahdanau:ICLR2015,Wu:Arxiv2016,Vaswani:NIPS2017}.
With the development of NMT,
prevailing NMT models become more and more complex with large numbers of parameters, which often require abundant corpora for effective training.
However, for translation tasks in most domains, 
domain-specific parallel sentences are often scarce.
If we only use domain-specific data to train the NMT model for such a domain, 
the performance of resulting model is usually unsatisfying.
Therefore,
NMT for low-resource domains becomes a challenge in its research and applications.

To deal with this issue,
many researchers have conducted studies on the domain adaptation for NMT, 
which can be classified into two general categories.
One is to transfer the rich-resource domain (out-of-domain) translation knowledge to benefit the low-resource (in-domain) NMT model. 
The other is to use the mixed-domain training corpus to construct a unified NMT model for all domains.
Here, we mainly focus on the first type of research, of which typical
methods include fine-tuning \cite{Luong:IWSLT2015,Zoph:EMNLP2016,Servan:Arxiv2016}, mixed fine-tuning \cite{Chu:ACL2017}, cost weighting \cite{Chen:FWNMT2017}, data selection \cite{Wang:ACL2017,Wang:EMNLP2017,zhang:NAACL2019} and so on.
The underlying assumption of these approaches is that in-domain and out-of-domain NMT models share the same parameter space or prior distributions, 
and the useful out-of-domain translation knowledge can be completely transferred to in-domain NMT model in a one-pass manner.
However, 
it is difficult to achieve this goal due to domain differences.
Particularly,
when the domain difference is significant,
such conventional brute-force transfer may be unsuccessful, 
facing the similar issue as the domain adaptation for other tasks \cite{Pan:IEEE2010}. 

In this paper, 
to tackle the above problem, 
we argue that corpora of different domains should be repeatedly utilized to fully distill domain-shared translation knowledge.
To this end, 
we propose a novel \textbf{I}terative \textbf{D}ual \textbf{D}omain \textbf{A}daptation (IDDA) framework for NMT.
Under this framework, 
we first train in-domain and out-of-domain NMT models using their own training corpora respectively, 
and then iteratively perform bidirectional translation knowledge transfer (from in-domain to out-of-domain and then vice versa).
In this way, 
both in-domain and out-of-domain NMT models are expected to constantly reinforce each other, 
which is likely to achieve better NMT domain adaptation.
Particularly,
we employ a \emph{knowledge distillation} \cite{Hinton:arXiv2015,Kim:2016} based approach to transfer translation knowledge.
During this process, 
the target-domain NMT model is first initialized with the source-domain NMT model, 
and then trained to fit its own training data and match the output of its previous best model simultaneously.
By doing so, 
the previously transferred translation knowledge can be effectively retained for better NMT domain adaptation.
Finally, 
we further extend the proposed framework to the scenario of multiple out-of-domain training corpora, 
where the above-mentioned bidirectional knowledge transfer is performed sequentially between the in-domain and each out-of-domain NMT models in the ascending order of their domain similarities.

The contributions of this work are summarized as follows:
\begin{itemize}
	\setlength{\itemsep}{0pt}
	\item We propose an iterative dual domain adaptation framework for NMT, which is applicable to many conventional domain transfer approaches, such as fine-tune, mixed fine-tune.
	Compared with previous approaches, our framework is able to better exploit domain-shared translation knowledge for NMT domain adaptation.
	
	\item We extend our framework to the setting of multiple out-of-domain training corpora, which is rarely studied in machine translation. 
    Moreover, we explicitly differentiate the contributions of different out-of-domain training corpora based on the domain-level similarity with in-domain training corpus.
	
	\item We provide empirical evaluations of the proposed framework on
	Chinese-English, German-English datasets for NMT domain adaptation. 
	Experimental results demonstrate the effectiveness of our framework.
	Moreover, 
	we deeply analyze impacts of various factors on our framework\footnote{We release code and results at https://github.com/DeepLearnXMU/IDDA.}.
	
% 	\item 
\end{itemize}

\section{Related Work}
Our work is obviously related to the research on transferring the out-of-domain translation knowledge into the in-domain NMT model.
In this aspect,
fine-tuning \cite{Luong:IWSLT2015,Zoph:EMNLP2016,Servan:Arxiv2016} is the most popular approach,
where
the NMT model is first trained using the out-of-domain training corpus, 
and then fine-tuned on the in-domain training corpus.
To avoid overfitting, 
\citet{Chu:ACL2017} blended in-domain with out-of-domain corpora to fine-tune the pre-trained model, 
and \citet{Freitag:Arxiv2016} combined the fine-tuned model with the baseline via ensemble method. 
Meanwhile, 
applying data weighting into NMT domain adaptation has attracted much attention.
\citet{Wang:ACL2017} and \citet{Wang:EMNLP2017} proposed several sentence and domain weighting methods with a dynamic weight learning strategy. 
\citet{zhang:NAACL2019} ranked unlabeled domain training samples based on their similarity
to in-domain data, and then adopts a probabilistic curriculum learning strategy during training.
\citet{Chen:FWNMT2017} applied the sentence-level cost weighting to refine the training of NMT model.
Recently, 
\citet{David:NAACL2018} introduced a weight to each hidden unit of out-of-domain model.
\citet{Chu:COLING2018} gave a comprehensive survey of the dominant domain
adaptation techniques for NMT.
\citet{Gu:NAACL2019} not only maintained a private encoder and a private decoder for each domain, but also introduced a common encoder and a common decoder shared by all domains.

\begin{figure*}[!t]
	\centering
	\includegraphics[width=1\linewidth]{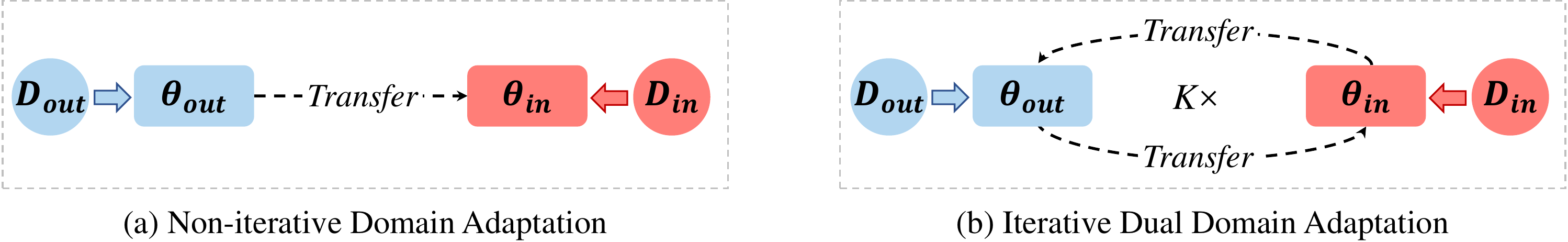}
	\caption{
		\label{OurModel}
		Traditional approach vs IDDA framework for one-to-one NMT domain adaptation.
		\textbf{\emph{$D_{out}$}}: out-of-domain training corpus, 
		\textbf{\emph{$D_{in}$}}: in-domain training corpus, 
		\textbf{\emph{$\theta_{out}$}}: out-of-domain NMT model, 
		\textbf{\emph{$\theta_{in}$}}: in-domain NMT model, 
		\textbf{\emph{K}} denotes the iteration number.
	}
\end{figure*}

\begin{algorithm*} [!t]
	%\small
	\caption{Iterative Dual Domain Adaptation for NMT}  
	\begin{algorithmic}[1] %每行显示行号  
		\State \textbf{Input:} Training corpora $\{D_{in},D_{out}\}$, development sets $\{D^{v}_{in}, D^{v}_{out}\}$, and the maximal iteration number $K$.
		\State \textbf{Output:} In-domain NMT model $\theta^*_{in}$.
		\State $\theta^{(0)}_{in}$ $\leftarrow$ \text{TrainModel}($D_{in}$), \ \ \ $\theta^{(0)}_{out}$ $\leftarrow$ \text{TrainModel}($D_{out}$)
		\State $\theta^*_{in}$ $\leftarrow$ $\theta^{(0)}_{in}$, \ \ \ $\theta^*_{out}$ $\leftarrow$ $\theta^{(0)}_{out}$
		\State \textbf{for} $k=1,2,...,K$ \textbf{do}
		\State \ \ \ \ $\theta^{(k)}_{out}$ $\leftarrow$
		\text{TransferModel}($\theta^{(k-1)}_{in}$, \ $D_{out}$, \ $\theta^{*}_{out}$)
		\State \ \ \ \ \textbf{if} \text{EvalModel}($D^v_{out}$, $\theta_{out}^{(k)}$) $>$ \text{EvalModel}($D^v_{out}$, $\theta^*_{out}$)
		\State \ \ \ \ \ \ \ \ \ \ \ \ $\theta^*_{out} \leftarrow \theta^{(k)}_{out}$
		\State \ \ \ \ \textbf{end if}
		\State \ \ \ \ $\theta^{(k)}_{in}$ $\leftarrow$
		\text{TransferModel}($\theta^{(k)}_{out}$, \ $D_{in}$, \ $\theta^{*}_{in}$)
		\State \ \ \ \ \textbf{if} \text{EvalModel}($D^v_{in}$, $\theta_{in}^{(k)}$) $>$ \text{EvalModel}($D^v_{in}$, $\theta^*_{in}$)
		\State \ \ \ \ \ \ \ \ \ \ \ \ $\theta^*_{in}$ $\leftarrow$ $\theta^{(k)}_{in}$
		\State \ \ \ \ \textbf{end if}
		\State \bf{end for}
	\end{algorithmic}\label{Algorithm1}
\end{algorithm*}

Significantly different from the above methods,
along with the studies of dual learning for NMT \cite{He:NIPS2016,Wang:AAAI2018,Zhang:AAAI2019}, 
we iteratively perform bidirectional translation knowledge transfer between in-domain and out-of-domain training corpora.
To the best of our knowledge, our work is the first attempt to explore such a dual learning based framework for NMT domain adaptation.
Furthermore, 
we extend our framework to the scenario of multiple out-of-domain corpora.
Particularly, 
we introduce knowledge distillation into the domain adaptation for NMT and experimental results demonstrate its effectiveness,
echoing its successful applications on many tasks,
such as speech recognition \cite{Hinton:arXiv2015} and natural language processing \cite{Kim:2016,Tan:2019}.

Besides, 
our work is also related to the studies of multi-domain NMT, 
which focus on building a unified NMT model trained on the mixed-domain training corpus for translation tasks in all domains \cite{Kobus:Arxiv2016,Tars:arXiv2018,Farajian:WMT2017,Pryzant:WMT2017,Sajjad:arXiv2017,Zeng:EMNLP2018, Bapna:NAACL2019}.
Although our framework is also able to refine out-of-domain NMT model, 
it is still significantly different from multi-domain NMT,
since only the performance of in-domain NMT model is considered.

Finally, 
note that similar to our work, 
\citet{Tan:2019} introduced knowledge distillation into multilingual NMT.
However, 
our work is still different from \cite{Tan:2019} in the following aspects: 
(1) \citet{Tan:2019} mainly focused on constructing a unified NMT model for multi-lingual translation task, 
while we aim at how to effectively transfer out-of-domain translation knowledge to in-domain NMT model;
(2)
Our translation knowledge transfer is bidirectional, 
while the procedure of knowledge distillation in \cite{Tan:2019} is unidirectional;
(3)  
When using knowledge distillation under our framework, 
we iteratively update teacher models for better domain adaptation.
In contrast, 
all language-specific teacher NMT models in \cite{Tan:2019} remain fixed.

\section{Iterative Dual Domain Adaptation Framework}
In this section, 
we first detailedly describe our proposed framework for conventional one-to-one NMT domain adaptation, 
and then extend this framework to the scenario of multiple out-of-domain corpora (many-to-one).

\begin{figure*}[!t]
	\centering
	\includegraphics[width=0.9\linewidth]{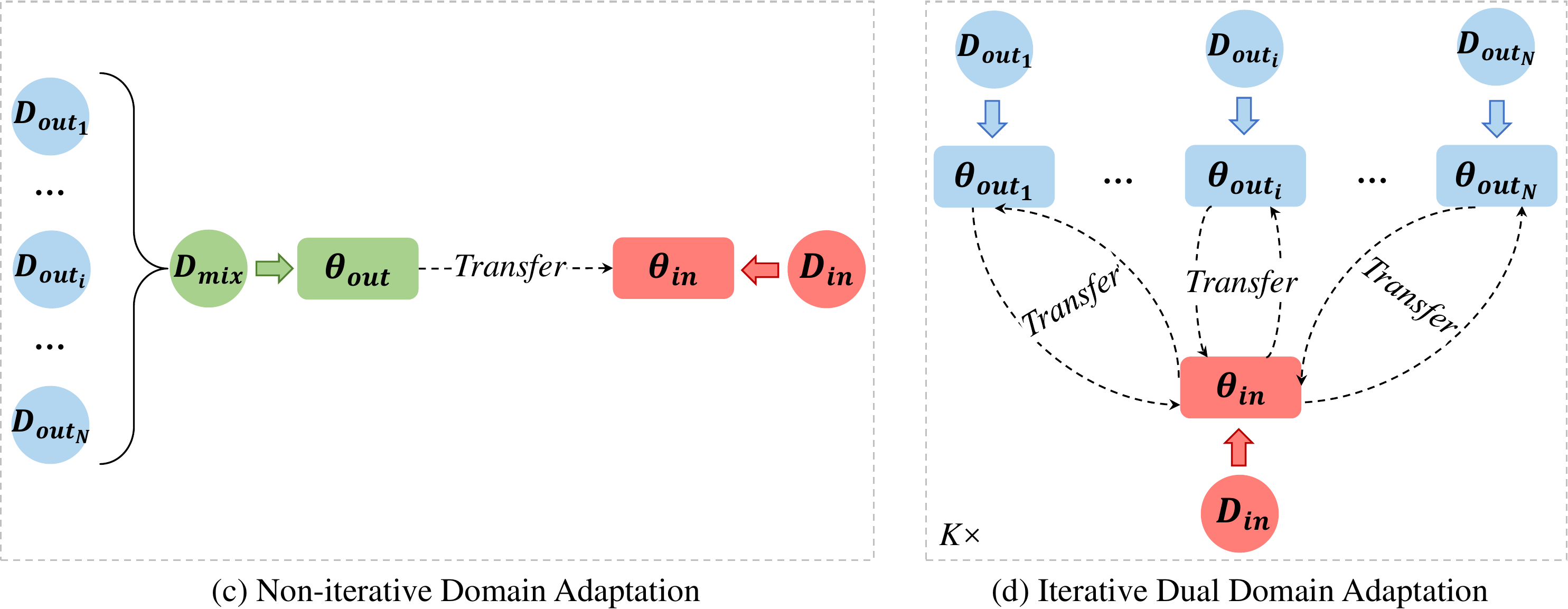}
	\caption{
		\label{OurModel2}
		Traditional approach vs IDDA framework for many-to-one NMT domain adaptation.
		\textbf{\emph{$D_{mix}$}}: a mixed out-of-domain training corpus. 
	}
\end{figure*}

\subsection{One-to-one Domain Adaptation}
As shown in Figure \ref{OurModel}(a), 
previous studies mainly focus on the one-pass translation knowledge transfer from one out-of-domain NMT model to the in-domain NMT model.
Unlike these studies,
we propose to conduct iterative dual domain adaptation for NMT,
of which framework is illustrated in Figure \ref{OurModel}(b).

To better describe our framework, 
we summarize the training procedure of our framework in Algorithm \ref{Algorithm1}.
Specifically,
we first individually train the initial in-domain and out-of-domain NMT models, respectively
denoted by $\theta^{(0)}_{in}$ and $\theta^{(0)}_{out}$, 
via minimizing the negative likelihood of their own training corpora $D_{in}$ and $D_{out}$ (\textbf{Line 3}):
\begin{equation}
\mathcal{L}^{(0)}_{in} = \sum_{(\mathbf{x},\mathbf{y})\in D_{in}}-log P(\mathbf{y}|\mathbf{x};\theta^{(0)}_{in}), 
\end{equation}
\begin{equation}
\mathcal{L}^{(0)}_{out} = \sum_{(\mathbf{x},\mathbf{y})\in D_{out}}-log P(\mathbf{y}|\mathbf{x};\theta^{(0)}_{out}).
\end{equation}
Then, 
we iteratively perform bidirectional translation knowledge transfer to update both in-domain and out-of-domain NMT models,
until the maximal iteration number $K$ is reached (\textbf{Lines 5-14}).
More specifically,
at the $k$-th iteration,
we first transfer the translation knowledge of the previous in-domain NMT model $\theta^{(k-1)}_{in}$ to the out-of-domain NMT model $\theta^{(k)}_{out}$ trained on $D_{out}$ (\textbf{Line 6}),
and then reversely transfer
the translation knowledge encoded by $\theta^{(k)}_{out}$ to the in-domain NMT model $\theta^{(k)}_{in}$ trained on $D_{in}$ (\textbf{Line 10}).
During this process,
we evaluate the new models $\theta^{(k)}_{in}$ and $\theta^{(k)}_{out}$ on their corresponding development sets, and then record the best model parameters as $\theta^{*}_{in}$ and $\theta^{*}_{out}$ (\textbf{Lines 7-9, 11-13}).

Obviously, 
during the above procedure, 
one of important steps is how to transfer the translation knowledge from one domain-specific NMT model to the other one.
However, 
if we directly employ conventional domain transfer approaches, 
such as fine-tuning, 
as the iterative dual domain adaptation proceeds,
the previously learned translation knowledge tends to be ignored.
To deal with this issue, we introduce knowledge distillation \cite{Kim:2016} to conduct the translation knowledge transfer.
Specifically, 
during the transfer process from $\theta^{(k)}_{out}$ to $\theta^{(k)}_{in}$,
we first initialize $\theta^{(k)}_{in}$ with parameters of $\theta^{(k)}_{out}$, 
and then train $\theta^{(k)}_{in}$ not only to match the references of $D_{in}$, 
but also to be consistent with probability outputs of the previous best in-domain NMT model $\theta_{in}^*$,
which is considered as the teacher model. 
To this end, 
we define the loss function as
\begin{align} \label{KL_loss}
\mathcal{L}^{(k)}_{in} =& \sum_{(\mathbf{x},\mathbf{y})\in D_{in}}[- (1-\lambda) \cdot log P(\mathbf{y}|\mathbf{x};\theta^{(k)}_{in}) + \notag \\ 
&\lambda\cdot \text{KL}(P(\mathbf{y}|\mathbf{x};\theta^{(k)}_{in})||P(\mathbf{y}|\mathbf{x};\theta^*_{in}))],
\end{align}
where $\lambda$ is the coefficient used to trade off these two loss terms, 
and it can be tuned on the development set. 
Notably, when $\lambda$=0, 
only the term of likelihood function affects the model training,
and thus our transfer approach degenerate into fine-tuning at each iteration.

In this way, 
we enable in-domain NMT model $\theta^{(k)}_{in}$ to not only retain the previously learned effective translation knowledge, 
but also fully absorb the useful translation knowledge from out-of-domain NMT model $\theta^{(k)}_{out}$.
Similarly, 
we employ the above method to transfer translation knowledge from $\theta^{(k-1)}_{in}$ to $\theta^{(k)}_{out}$ using out-of-domain corpus $D_{out}$ and the previous best out-of-domain model $\theta^{*}_{out}$. 
Due to the space limitation, 
we omit the specific description of this procedure.

\subsection{Many-to-one Domain Adaptation} \label{Sec_many_to_one}
Usually, 
in practical applications,
there exist multiple available out-of-domain training corpora simultaneously.
%out-of-domain corpora at the same time. 
As shown in Figure \ref{OurModel2}(a), 
previous studies usually mix them into one out-of-domain corpus, which is applicable for the conventional one-to-one NMT domain adaptation.
However, various out-of-domain corpora are semantically related to in-domain corpus to different degrees, 
and thus intuitively,
it is difficult to adequately play their roles without distinguishing them.

To address this issue, 
we extend the proposed framework to many-to-one NMT domain adaptation.
Our extended framework is illustrated in Figure \ref{OurModel2}(b).
Given an in-domain corpus and $N$ out-of-domain corpora,
we first measure the semantic distance between each out-of-domain corpus and the in-domain corpus using the proxy $A$-distance $\hat{d}_A$=$2(1-2\epsilon)$ \cite{Ganin:MLR2015,Pryzant:WMT2017}, where the $\epsilon$ is the generalization error of a linear bag-of-words SVM classifier trained to discriminate between the two domains.
Then, 
%we determine the transfer order $\{Out_1,Out_2,...,Out_N\}$ of these datasets,
we determine the transfer order of these out-of-domain NMT models as 
$\{\theta_{out_1}, \theta_{out_2},...\theta_{out_N}\}$ ,
according to distances of their own training corpora
%their distances 
to the in-domain corpus in a decreasing order.
The reason behind this step is the translation knowledge of 
%pre ordinal 
previously transferred out-of-domain NMT models will be partially forgotten during the continuous transfer.
By setting transfer order according to their $\hat{d}_A$ values in a decreasing order, 
we enable the in-domain NMT model to fully preserve the translation knowledge transferred from the most relevant out-of-domain NMT model.
%the most relevant translation knowledge in the adapted NMT model as far as possible.
Finally,
we sequentially perform bidirectional knowledge transfer between the in-domain and each out-of-domain models, where this process will be repeated for $K$ iterations.

\section{Experiments}
To verify the effectiveness of our framework, 
we first conducted one-to-one domain adaptation experiments on Chinese-English translation, 
where we further investigated impacts of various factors on our framework. 
Then, we carried out two-to-one domain adaptation experiments on English-German translation, 
so as to demonstrate the generality of our framework on different language pairs and multiple out-of-domain corpora.

\subsection{Setup}
\textbf{Datasets.}
In the Chinese-English translation task,
our in-domain training corpus is from IWSLT2015 dataset consisting of 210K \emph{TED Talk} sentence pairs, 
and 
the out-of-domain training corpus contains 1.12M LDC sentence pairs related to \emph{News} domain.
For these two domains,
we chose IWSLT dev2010 and NIST 2002 dataset as development sets.
Finally, we used IWSLT tst2010, tst2011 and tst2012 as in-domain test sets.
Particularly, in order to verify whether our framework can enable NMT models of two domains to benefit each other,
we also tested the performance of out-domain NMT model on NIST 2003, 2004, 2005, 2006 datasets.

For the English-German translation task, 
our training corpora totally include one in-domain dataset:
200K \emph{TED Talk} sentence pairs provided by IWSLT2015,
and 
two out-of-domain datasets: 
500K sentence pairs (\emph{News} topic) extracted from WMT2014 corpus, 
and 
500K sentence pairs (\emph{Medical topic}) that are sampled from OPUS EMEA corpus\footnote{http://opus.nlpl.eu/}.
As for development sets, 
we chose IWSLT tst2012, WMT tst2012 and 1K sampled sentence pairs of OPUS EMEA corpus, respectively. 
In addition, IWSLT tst2013, tst2014 were used as in-domain test sets, WMT news-test2014 (News topic) and 1K sampled sentence pairs of OPUS EMEA corpus were used as two out-of-domain test sets. 

We first employed \emph{Stanford Segmenter}\footnote{https://nlp.stanford.edu/} to conduct word segmentation on Chinese sentences and \emph{MOSES script}\footnote{http://www.statmt.org/moses/} to tokenize English and German sentences. 
Then, we limited the length of sentences to 50 words in the training stage.
Besides, we employed \emph{Byte Pair Encoding} \cite{Sennrich:ACL2016} to split words into subwords and
set the vocabulary size for both Chinese-English and English-German as 32,000. 
We evaluated the translation quality with BLEU scores \cite{Papineni:ACL2002} as calculated by \texttt{multi-bleu.perl} script .

\noindent\textbf{Settings.}
We chose Transformer \cite{Vaswani:NIPS2017} as our NMT model, which exhibits excellent performance due to its flexibility in parallel computation and long-range dependency modeling.
We followed \citet{Vaswani:NIPS2017} to set the configurations.
The dimensionality of all input and output layers is 512, and that of FFN layer is 2048. We employed 8 parallel attention heads in both encoder and decoder.
Parameter optimization was performed using stochastic gradient descent, where \emph{Adam} \cite{Kingma:ICLR2015} was used to automatically adjust the learning rate of each parameter. 
We batched sentence pairs by approximated length, and limited input and output tokens per batch to 25000 tokens. 
%During training, 
%we used label smoothing with value $0.1$, attention dropout and residual dropout with a rate of $0.1$. 
As for decoding 
we employed beam search algorithm and set the beam size as 4.
Besides, we set the distillation coefficient $\lambda$ as $0.4$.

\noindent\textbf{Contrast Models.}
We compared our framework with the following models, namely:
\begin{itemize}
	\setlength{\itemsep}{0pt}	
	\item {\bf Single}
	A reimplemented Transformer only trained on a single domain-specific (in/out) training corpus.
	
	\item{\bf Mix}
	A reimplemented Transformer trained on the mix of in-domain and out-of-domain training corpora.
	%mix-domain (in+out) training corpus.
	\item {\bf Fine-tuning (FT)} \cite{Luong:IWSLT2015}.
	It first trains the NMT model on out-of-domain training corpus
	%general domain training corpus 
	and then fine-tunes it using in-domain training corpus.
	
	\item{\bf Mixed Fine-tuning (MFT)} \cite{Chu:ACL2017}.
	It also first trains the NMT model on out-of-domain training corpus, and then fine-tunes it  
	%Different from the previous model,
	%it fine-tunes the pre-trained out-of-domain NMT model 
	using both out-of-domain and over-sampling in-domain training corpora.
	%It first trained NMT model using general domain training dataset, and then fine-tuning it using the mixed training corpus composed of general domain dataset and over-sampling target domain dataset.
	
	\item{\bf Knowledge Distillation (KD)} \cite{Kim:2016}.
	Using this method,
	we first train a out-of-domain and an in-domain NMT models using their own training corpus, respectively.
	Then, we use the in-domain training corpus to fine-tune the out-of-domain NMT model, supervised by the in-domain NMT model.
	%It first trained general domain NMT model and target domain NMT model, respectively. And then trained general domain NMT model teached by target domain NMT model on target domain dataset.
\end{itemize}

Besides, we reported the performance of some recently proposed multi-domain NMT models.

\begin{table*}[!t]
	\centering
	\small
	\begin{tabular}{l|ccccc|ccccc}
		\hline
		\multirow{2}{*}{\textbf{Model}} & \multicolumn{5}{c|}{TED Talk (\textbf{In-domain})}  & \multicolumn{5}{c}{News (\textbf{Out-of-domain})} \\
		\cline{2-11}
		& Tst10 & Tst11 & Tst12 & Tst13  & \textsc{Ave.} 
		& Nist03 & Nist04 & Nist05 & Nist06
		& \textsc{Ave.} \\
		\hline
		\hline
		\multicolumn{11}{c}{\emph{Cross-domain Transfer Methods}} \\
		\hline
		%\multicolumn{6}{l}{\textbf{(Out $\rightarrow$ In) domain transfer Methods}} \\
		Single     & 15.82 & 20.80 & 17.77 & 18.33 & 18.18  & 45.38 & 45.93 & 42.80 & 42.70 & 44.20 \\
		Mix        & 16.46 & 20.85 & 19.13 & 19.87 & 19.08 & 44.87 & 45.71 & 42.24 & 42.02 & 43.71 \\
		FT  & 16.77 & 21.16 & 19.31 & 20.53 & 19.44 & --- & --- & --- & --- & --- \\
		MFT & 17.19 & 22.02 & 20.09 & 21.05 & 20.08 & --- & --- & --- & --- & --- \\
		KD  & 17.62 & 21.88 & 19.97 & 20.43 & 19.98 & --- & --- & --- & --- & --- \\
		\hline
		\multicolumn{11}{c}{\emph{Multi-domain NMT Methods}} \\
		\hline	
		DC  & 17.23 & 22.10 & 19.68 & 20.58 & 19.90 & 46.03 & 46.62 & 44.39 & 43.82 & 45.21 \\
		%MFT & 17.19 & 22.02 & 20.09 & 21.05 & 20.08 & 45.51 & 46.03 & 44.05 & 43.39 & 44.74 \\		
		DM  & 16.45 & 21.35 & 18.77 & 20.27 & 19.21 & 45.12 & 45.83 & 42.77 & 42.59 & 44.08 \\
		WDCD & 17.32 & 22.23 & 20.02 & 21.10 & 20.17 & 46.33 & 46.36 & 44.62 & 43.80 &  45.27 \\
		\hline
		\multicolumn{11}{c}{\emph{IDDA Framework}} \\
		\hline
		% & 16.97 & 21.69 & 20.13 & 20.86
		%MFT \ding{192}$\leftarrow$\ding{193} & xx & xx & xx & xx & xx & - & - & - & - & - \\
		%KD  \ding{192}$\leftarrow$\ding{193} & 17.62 & 21.88 & 19.97 & 20.43 & 19.98 & -- & -- & -- & -- & -- \\	
		%Adversarial     & 16.97 & 21.69 & 19.53 & 20.56 & 45.42 & 46.37 & 43.69 & 43.11 \\
		%Target Token    & 17.14 & 21.99 & 19.70 & 20.70 & 46.58 & 46.59 & 44.06 & 43.73 \\		
		IDDA($\lambda$=0)  & 18.00 & 22.71 & 20.36 & \textbf{21.82} & 20.72 & 45.91 & 45.84 & 43.61 & 42.17 & 44.46 \\
		IDDA  & \textbf{18.36} & \textbf{23.14} & \textbf{20.78} & 21.79 & \textbf{21.02}$^\dag$ & \textbf{47.17} & \textbf{47.44} & \textbf{45.38} & \textbf{44.04} & \textbf{46.01}$^\dag$ \\
		\hline
	\end{tabular}
	\caption{\label{Table-c2eExp}
		Experimental results on the Chinese-English translation task. 
		\textbf{\dag} indicates statistically significantly better than ($\rho<$0.01) the result of \emph{WDCD}.
	}
\end{table*}

\begin{itemize}	
	\setlength{\itemsep}{0pt}
	\item {\bf Domain Control (DC)} \cite{Kobus:Arxiv2016}.
	It is also based on the mix-domain NMT model.
	However,
	it adds an additional domain tag to each source sentence, incorporating domain information into source annotations.
	
	\item {\bf Discriminative Mixing (DM)} \cite{Pryzant:WMT2017}.
	It jointly trains NMT with domain classification via multitask learning. Please note that it performs the best among three approaches proposed by Pryzant et al., \shortcite{Pryzant:WMT2017}.
	
	\item {\bf Word-level Domain Context Discrimination (WDCD)} \cite{Zeng:EMNLP2018}.
	It discriminates the source-side word-level domain specific and domain-shared contexts for multi-domain NMT by jointly modeling NMT and domain classifications.
\end{itemize}

\subsection{Results on Chinese-English Translation}

\subsubsection{Effect of Iteration Number $K$}
%Obviously, 
The iteration number $K$ is a crucial hyper-parameter that directly determines the amount of the transferred translation knowledge under our framework.
Therefore, 
we first inspected its impacts on the development sets. 
To this end, 
we varied $K$ from 0 to 7 with an increment of 1 in each step, 
where our framework degrades to \emph{Single} when $K$=0.

\begin{figure}[!t]
	\centering
	\includegraphics[width=1.0\linewidth]{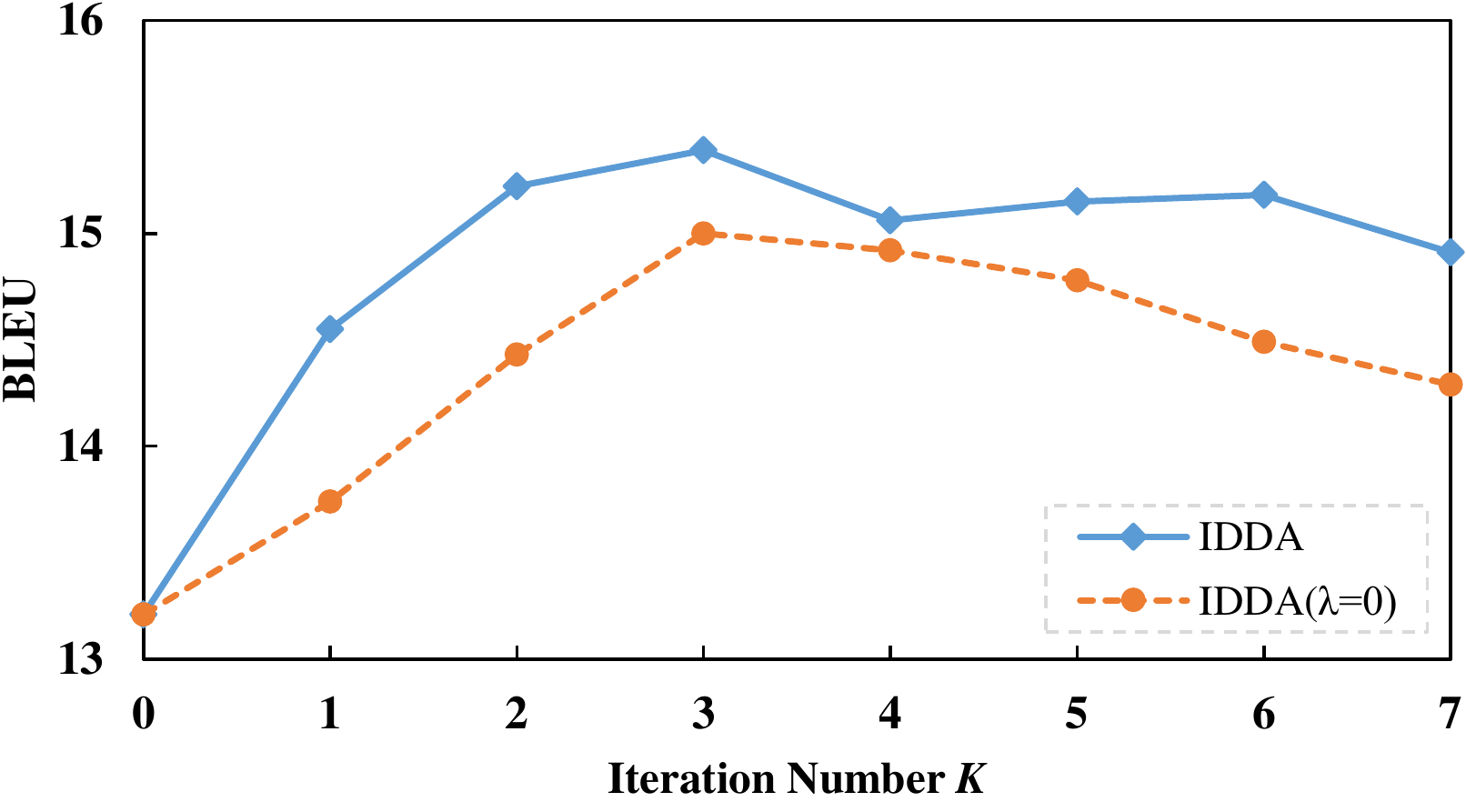}
	\caption{
		\label{Fig-Iter-IWSLT}
		Effect of iteration number ($K$) on the Chinese-English in-domain development set.
	}
\end{figure}	

Figure \ref{Fig-Iter-IWSLT} provides the experimental results using different $K$s. 
%of the domain adaptation from \emph{News} domain to \emph{TED Talk} domain. 
We can observe that both \emph{IDDA($\lambda$=0)} and \emph{IDDA} achieve the best performance at the $3$-th iteration, respectively.
Therefore, 
we directly used $K$=3 in all subsequent experiments.	

\subsubsection{Overall Performance} \label{Ch_En_Overal_Result}

Table \ref{Table-c2eExp} shows the overall experimental results. 
On all test sets, 
our framework significantly outperforms other contrast models. 
Furthermore, 
we reach the following conclusions:

First, on the in-domain test sets, both \emph{IDDA($\lambda$=0)} and \emph{IDDA} surpass
\emph{Single}, \emph{Mix}, \emph{FT}, \emph{MFT} and \emph{KD}, most of which are commonly used in the domain adaptation for NMT.
%our \emph{IDDA($\lambda$=0)} still exhibit better performance
%on all in-domain test sets, and achieves comparabel BLEU scores with contrst models on out-of-domain datasets.
This confirms the difficulty in completely one-pass transferring the useful out-of-domain translation knowledge to the in-domain NMT model.
% through one time.
Moreover, the in-domain NMT model benefits from multiple-pass knowledge transfers
%multiple knowledge transfer 
%with the out-of-domain NMT model 
under our framework.
%Thus, the transfer
%process should be iterative and incremental.

Second, 
compared with \emph{DC}, \emph{DM} and \emph{WDCD} that are proposed for multi-domain NMT, both \emph{IDDA($\lambda$=0)}
and \emph{IDDA} still exhibit better performance on the in-domain test sets.
The underlying reason is that
these multi-domain models discriminate domain-specific and domain-shared information in encoder,
however, 
their shared decoder are inadequate to effectively preserve domain-related text style and idioms.
In contrast,
our framework is adept at preserving these information since we construct an individual NMT model for each domain.

Third, 
\emph{IDDA} achieves better performance than \emph{IDDA($\lambda$=0)}, demonstrating the importance of retaining previously learned translation knowledge.
Surprisingly, \emph{IDDA} significantly outperforms \emph{IDDA($\lambda$=0)} on out-of-domain data sets.
We conjecture that during the process of knowledge distillation, by assigning non-zero probabilities to multiple words, 
the output distribution of teacher model is more smooth, 
leading to smaller variance in gradients \cite{Hinton:arXiv2015}.
Consequently, 
the out-of-domain NMT model becomes more robust by iteratively absorbing the translation knowledge from the best out-of-domain model.

Finally, note that even on the out-of-domain test
sets, \emph{IDDA} still has better performance than all
listed contrast models in the subsequent experimental analyses. This result demonstrates
the advantage of dual domain adaptation under our
framework.
%Finally, note that even on the out-of-domain test sets,
%\emph{IDDA} still has better performance than all listed contrast models.
%This result demonstrates the advantage of dual domain adaptation under our framework.

According to the reported performance of our framework shown in Table \ref{Table-c2eExp},
we only considered \emph{IDDA} in all subsequent experiments.
Besides,
we only chose \emph{MFT}, \emph{KD}, and \emph{WDCD} as typical contrast models.
This is because 
\emph{KD} is the basic domain adaption approach of our framework,
\emph{MFT} and \emph{WDCD} are the best domain adaptation method and multi-domain NMT model for comparison, respectively.

\subsubsection{Results on Source Sentences with Different Lengths}
%\subsubsection{Effects on Sentence Lengths.}
\begin{figure}[!t]
	\centering
	\small
	\includegraphics[width=1.0\linewidth]{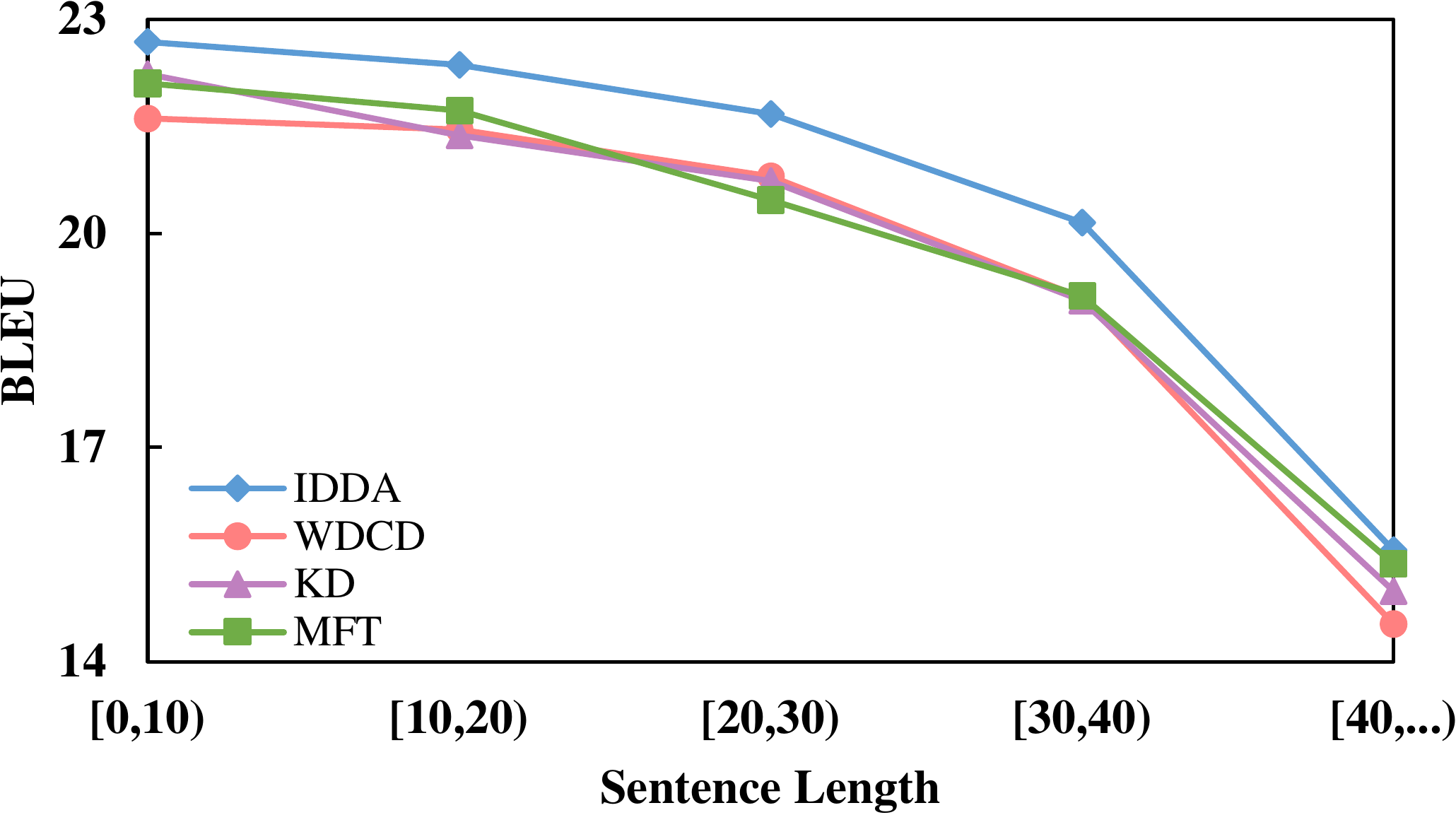}
	\caption{
		\label{Figure_Length_IWSLT}
		BLEU scores on different IWSLT test sets divided according to source sentence lengths.
	}
\end{figure}

Following previous work \cite{Bahdanau:ICLR2015},
we divided IWSLT test sets into different groups based on the lengths of source sentences 
and then investigated the performance of various models.

Figure \ref{Figure_Length_IWSLT} illustrates the results.
We observe that 
our framework also achieves the best performance in all groups, 
although the performances of all models degrade with the increase of the length of source sentences.

\subsubsection{Effect of Out-of-domain Corpus Size}
%In previous studies,
%researchers often assume that the size of out-of-domain corpus is larger than that of in-domain corpus.
%In this group of experiments,
%we investigated the effectiveness of our proposed framework using different sizes of out-of-domain corpora:
%50K, 200K and 1.12M, respectively.
In this group of experiments, we investigated the impacts of out-of-domain corpus size on our proposed framework.
Specifically, we inspected the results of our framework using  different sizes of out-of-domain corpora: 50K, 200K and 1.12M, respectively

\begin{figure}[!t]
	\centering
	\includegraphics[width=1.0\linewidth]{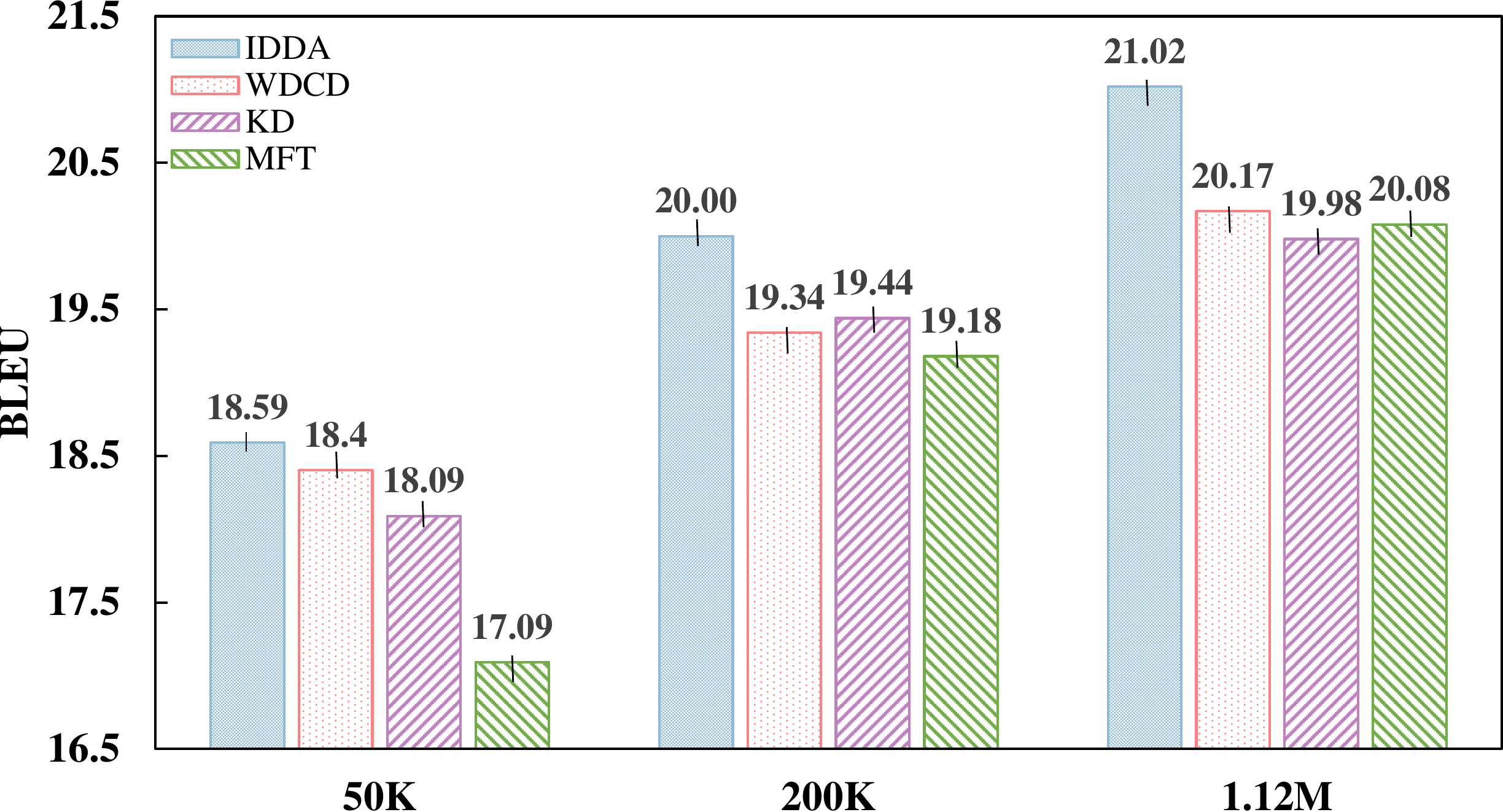}
	\caption{
		\label{Figure_Corpus_Size}
		Experimental results with different sizes of out-of-domain corpora.
	}
\end{figure}

Figure \ref{Figure_Corpus_Size} shows the comparison results on the average BLEU scores of all IWSLT test sets.
No matter how large out-of-domain data is used,
\emph{IDDA} always achieves better performance than other contrast models,
demonstrating the effectiveness and generality of our framework.
Specially, 
\emph{IDDA} with 200K out-of-domain corpus is comparable to \emph{KD} with 1.12M corpus. 
From this result, 
we confirm again that our framework is able to better exploit the complementary information between domains than \emph{KD}.

\subsubsection{Effects of Dual Domain Adaptation and Updating Teacher Models}
\begin{table}[!t]
	\centering
	\small
	\begin{tabular}{c|c}
		\hline
		\textbf{Model} & \textsc{Ave.}  \\
		\hline
		IDDA-unidir & 20.43 \\
		IDDA-fixTea & 20.60 \\
		IDDA & 21.02 \\
		\hline
	\end{tabular}
	\caption{\label{Result_Dynamic_vs_Fixed}
		Experimental results of comparing IDDA with its two variants.
	}
\end{table}
Two highlights of our framework consist of the usage of bidirectional translation knowledge transfer and continuous updating teacher models $\theta^*_{out}$ and $\theta^*_{in}$ (See Line 6, 10 of Algorithm \ref{Algorithm1}).
To inspect their effects on our framework,
we compared our framework with its two variants:
(1) \emph{IDDA-unidir}, where we only iteratively transfer out-of-domain translation knowledge to the in-domain NMT model;
(2) \emph{IDDA-fixTea}, where teacher models are fixed as the initial out-of-domain and in-domain NMT models, respectively.

The results are displayed in Table \ref{Result_Dynamic_vs_Fixed}.
We can see that our framework exhibits better performance than its two variants, 
which demonstrates that dual domain adaptation enables NMT models of two domains to benefit from each other, 
and updating teacher models is more helpful to retain useful translation knowledge.
%Since our framework exhibits better performance than its two variants.
%we confirm that dual domain adaptation enables NMT models of two domains to benefit each other, and updating Teacher models is more helpful to retain useful translation knowledge.	

\subsubsection{Case Study}
\begin{table}[!t]
	\small
	\begin{center}
		{
			
			\begin{tabular}{c<{\centering}|l}
				\hline
				\textbf{Model} & \ \ \ \ \ \ \ \ \ \ \ \ \ \ \ \ \ \ \ \ \textbf{Translation} \\
				\hline
				%\hline
				\multirow{2}{*}{Src} &zh\`{e} sh\`{i} d\`{i} y\={i} zh\v{o}ng zh\'{i}l\`{i} x\'ingz\v{o}u de \\
				& l\'{i}ngzh\v{a}ngl\`{e}i d\`{o}ngw\`{u}\\
				\hline
				Ref  & that was the first upright primate \\
				\hline
				MFT  & this is the first \emph{\color {red} animal} to walk upright\\
				\hline
				KD & this is the first \emph{\color {red} growing} primate \\
				\hline
				WDCD  & this is the first primate \emph{\color {red} walking around}\\
				\hline
				IDDA-1 & this is the first \emph{\color {orange} upright - walking primate}\\
				\hline
				IDDA-2 & this is the first \emph{\color {orange} upright - walking primate}\\
				\hline
				IDDA-3 & this is the first \emph{\color {blue} primates walking upright}\\
				\hline
				IDDA-4 & this is the first \emph{\color {blue} upright primate}\\
				\hline
				IDDA-5 & this is the first \emph{\color {blue} upright primate}\\
				\hline
				IDDA-6 & this is the first \emph{\color {blue} upright primate}\\
				\hline
				IDDA-7 & this is the first \emph{\color {blue}upright primate}\\
				\hline
			\end{tabular}
		}
	\end{center}
	\caption{
		\label{Exp-ChiEng-Tran}
		Translation examples of different NMT models.
		\textbf{Src}: source sentence and \textbf{Ref}: target reference. \emph{IDDA-k} represents the in-domain NMT model at the $k$-th iteration using our framework.
	}
\end{table}

Table \ref{Exp-ChiEng-Tran} displays the 1-best translations of a sampled test sentence generated by \emph{MFT}, \emph{KD}, \emph{WDCD}, and \emph{IDDA} at different iterations.
Inspecting this example provides the insight into the advantage of our proposed framework to some extent.
Specifically, 
we observe that \emph{MFT}, \emph{KD}, \emph{WDCD} are unable to correctly understand the meaning of ``zh\'{i}l\`{i} x\'ingz\v{o}u de l\'{i}ngzh\v{a}ngl\`{e}i d\`{o}ngw\`{u}'' 
and thus generate incorrect or incomplete translations,
while \emph{IDDA} successfully corrects these errors by gradually absorbing transferred translation knowledge.

\subsection{Results on English-German Translation}
\subsubsection{Overall Performance}

\begin{table*}[!t]
	\centering
	\small
	\begin{tabular}{l|ccc|c|c}
		\hline
		\multirow{3}{*}{\textbf{Model}} & \multicolumn{3}{c|}{\textbf{In-domain}}  
		& \textbf{Out-of-domain1} & \textbf{Out-of-domain2} \\
		& \multicolumn{3}{c|}{TED Talk}  & News & Medical \\
		\cline{2-6}
		& IWSLT2013 & IWSLT2014 & \textsc{Ave.} & WMT14 & EMEA \\
		\hline
		\hline
		\multicolumn{6}{c}{\emph{Cross-domain Transfer Methods}} \\
		\hline
		%\hline
		% & 21.85
		Single     & 29.76 & 25.99 & 27.88 & 20.54 & 51.11 \\
		% & 25.00 
		Mix        & 31.45 & 27.03 & 29.24 & 21.17 & 50.60  \\
		FT       & 30.54 & 27.02 & 28.78 & --- & --- \\
		%& 25.33
		MFT     & 31.86 & 27.49 & 29.67 & --- & --- \\
		% & 23.68
		KD  & 31.33 & 27.96 & 29.64 & --- & --- \\
		\hline
		\multicolumn{6}{c}{\emph{Multi-domain NMT Methods}} \\
		\hline
		% & 25.16
		DC      & 31.13 & 28.02 & 29.57 & 21.61 & 52.25 \\
		%& 25.26
		DM      & 31.57 & 27.60 & 29.58 & 21.75 & 52.60 \\
		WDCD   & 31.87 & 27.82 &  29.84 & 21.86 & 52.84 \\
		\hline
		\multicolumn{6}{c}{\emph{IDDA Framework}} \\
		\hline
		%& 23.25
		IDDA($\lambda$=0) & 32.11 & 28.10 & 30.11 & 22.01 & 52.07 \\
		IDDA & \textbf{32.93} & \textbf{28.88} & \textbf{30.91}$^*$ & \textbf{22.17}$^\dag$ & \textbf{53.39}$^\dag$ \\
		\hline
	\end{tabular}
	\caption{\label{English_German}
		Experimental results of the English-German translation task.
		\textbf{*} indicates statistically significantly better than ($\rho<$0.05) the result of \emph{WDCD}.
	}
\end{table*}

We first calculated the distance between the in-domain and each out-of-domain corpora: $\hat{d}_A$(\emph{Ted Talk}, \emph{News}) = 0.92 and $\hat{d}_A$(\emph{Ted Talk}, \emph{Medical}) = 1.92.
Obviously, the \emph{News} domain is more relevant to \emph{TED Talk} domain than \emph{Medical} domain, 
and thus we determined the final transfer order as \{$\theta_{\rm out_{medical}}$, $\theta_{\rm out_{news}}$\}
%\{$Out_1$=\emph{Medical}, $Out_2$=\emph{News}\} 
for this task.
Then, as implemented in the previous Chinese-English experiments,
we determined the optimal $K$=2 on the development set.

Table \ref{English_German} shows experimental results.
%the results of English-German translation task.
Similar to the previously reported experiment results, 
our framework still obtains the best performance among all models, 
which verifies the effectiveness of our framework on many-to-one domain adaptation for NMT.

As described above,
we have two careful designs for many-to-one NMT domain adaptation:
(1) We distinguish different out-of-domain corpora, and then iteratively
perform bidirectional translation knowledge transfer
% perform dual domain adaptation 
between in-domain and each out-of-domain NMT models.
% the in-domain model and each out-of-domain model.
(2) We determine the transfer order according to the
semantic distance between each out-of-domain and in-domain training corpora.
Here, we carried out two groups of experiments to investigate their impacts on our framework.
In the first group of experiments,
we first combined all out-of-domain training corpora into a mixed corpus, 
and then applied our framework to establish the in-domain NMT model.
In the second group of experiments,
we employed our framework in different transfer orders to perform domain adaptation.

Table \ref{Result_Mix} shows the final experimental results,
which are in line with our expectations and verify the validity of our designs.

\begin{table}[!t]
	\centering
	\small
	\begin{tabular}{c|c|c}
		\hline
		\textbf{Model} & \textbf{Transfer Order} & \textsc{Ave.}  \\
		\hline
		IDDA-mix & ------ & 30.17 \\
		IDDA & \{$\theta_{\rm out_{news}}$, $\theta_{\rm out_{medical}}$\} & 30.51 \\
		IDDA & \{$\theta_{\rm out_{medical}}$, $\theta_{\rm out_{news}}$\} & 30.91 \\
		\hline
	\end{tabular}
	\caption{\label{Result_Mix}
		Experimental results of IDDA using different configurations.
	}
\end{table}

\section{Conclusion}
In this paper, 
we have proposed an iterative dual domain adaptation framework for NMT, 
which continuously fully exploits the mutual complementarity between in-domain and out-domain corpora for translation knowledge transfer. 
Experimental results and in-depth analyses on 
translation tasks of two language pairs 
strongly demonstrate the effectiveness of our framework.

In the future,
we plan to extend our framework to multi-domain NMT.
Besides, 
how to leverage monolingual sentences of different domains to refine our proposed framework.
Finally, 
we will apply our framework into other translation models \cite{Bahdanau:ICLR2015,Su:TASLP2018,Song:TACL2019}, 
so as to verify the generality of our framework.

%\subsubsection{Acknowledgments.}
\section*{Acknowledgments}
The authors were supported by National Natural Science Foundation of China (No. 61672440),
Beijing Advanced Innovation Center for Language Resources, 
NSF Award (No. 1704337),
the Fundamental Research Funds for the Central Universities (Grant No. ZK1024),
and Scientific Research Project of National Language Committee of China (Grant No. YB135-49). 
We also thank the reviewers for their insightful comments
	
\bibliography{emnlp-ijcnlp-2019}

\begin{thebibliography}{35}
\expandafter\ifx\csname natexlab\endcsname\relax\def\natexlab#1{#1}\fi

\bibitem[{Bahdanau et~al.(2015)Bahdanau, Cho, and Bengio}]{Bahdanau:ICLR2015}
Dzmitry Bahdanau, Kyunghyun Cho, and Yoshua Bengio. 2015.
\newblock Neural machine translation by jointly learning to align and
  translate.
\newblock In \emph{Proc. of ICLR 2015}.

\bibitem[{Bapna and Firat(2019)}]{Bapna:NAACL2019}
Ankur Bapna and Orhan Firat. 2019.
\newblock Non-parametric adaptation for neural machine translation.
\newblock In \emph{Proc. of NAACL 2019}.

\bibitem[{Chen et~al.(2017)Chen, Cherry, Foster, and Larkin}]{Chen:FWNMT2017}
Boxing Chen, Colin Cherry, George Foster, and Samuel Larkin. 2017.
\newblock Cost weighting for neural machine translation domain adaptation.
\newblock In \emph{Proc. of WMT 2018}.

\bibitem[{Chu et~al.(2017)Chu, Dabre, and Kurohashi}]{Chu:ACL2017}
Chenhui Chu, Raj Dabre, and Sadao Kurohashi. 2017.
\newblock An empirical comparison of domain adaptation methods for neural
  machine translation.
\newblock In \emph{Proc. of ACL 2017}.

\bibitem[{Chu and Wang(2018)}]{Chu:COLING2018}
Chenhui Chu and Rui Wang. 2018.
\newblock A survey of domain adaptation for neural machine translation.
\newblock In \emph{Proc. of COLING 2018}.

\bibitem[{Farajian et~al.(2017)Farajian, Turchi, Negri, and
  Federico}]{Farajian:WMT2017}
M.~Amin Farajian, Marco Turchi, Matteo Negri, and Marcello Federico. 2017.
\newblock Multi-domain neural machine translation through unsupervised
  adaptation.
\newblock In \emph{Proc. of WMT 2017}.

\bibitem[{Freitag and Al-Onaizan(2016)}]{Freitag:Arxiv2016}
Markus Freitag and Yaser Al-Onaizan. 2016.
\newblock Fast domain adaptation for neural machine translation.
\newblock \emph{CoRR abs/1612.06897.}

\bibitem[{Ganin et~al.(2015)Ganin, Ustinova, Ajakan, Germain, Larochelle,
  Laviolette, Marchand, and Lempitsky}]{Ganin:MLR2015}
Yaroslav Ganin, Evgeniya Ustinova, Hana Ajakan, Pascal Germain, Hugo
  Larochelle, Fran{\c{c}}ois Laviolette, Mario Marchand, and Victor~S.
  Lempitsky. 2015.
\newblock Domain-adversarial training of neural networks.
\newblock \emph{Machine Learning Research}, 17.

\bibitem[{Gu et~al.(2019)Gu, Feng, and Liu}]{Gu:NAACL2019}
Shuhao Gu, Yang Feng, and Qun Liu. 2019.
\newblock Improving domain adaptation translation with domain invariant and
  specific information.
\newblock In \emph{Proc. of NAACL 2019}.

\bibitem[{He et~al.(2016)He, Xia, Qin, Wang, Yu, Liu, and Ma}]{He:NIPS2016}
Di~He, Yingce Xia, Tao Qin, Liwei Wang, Nenghai Yu, Tie{-}Yan Liu, and
  Wei{-}Ying Ma. 2016.
\newblock Dual learning for machine translation.
\newblock In \emph{Proc. of NIPS 2016}.

\bibitem[{Hinton et~al.(2015)Hinton, Vinyals, and Dean}]{Hinton:arXiv2015}
Geoffrey~E. Hinton, Oriol Vinyals, and Jeffrey Dean. 2015.
\newblock Distilling the knowledge in a neural network.
\newblock \emph{CoRR abs/1503.02531}.

\bibitem[{Kim and Rush(2016)}]{Kim:2016}
Yoon Kim and Alexander~M. Rush. 2016.
\newblock Sequence-level knowledge distillation.
\newblock In \emph{Proc. of EMNLP 2016}.

\bibitem[{Kingma and Ba(2015)}]{Kingma:ICLR2015}
Diederik~P. Kingma and Jimmy~Lei Ba. 2015.
\newblock Adam: A method for stochastic optimization.
\newblock In \emph{Proc. of ICLR 2015}.

\bibitem[{Kobus et~al.(2016)Kobus, Crego, and Senellart}]{Kobus:Arxiv2016}
Catherine Kobus, Josep Crego, and Jean Senellart. 2016.
\newblock Domain control for neural machine translation.
\newblock \emph{CoRR abs/1612.06140.}

\bibitem[{Luong and Manning(2015)}]{Luong:IWSLT2015}
Minh-Thang Luong and Christopher~D Manning. 2015.
\newblock Stanford neural machine translation systems for spoken language
  domains.
\newblock In \emph{Proc. of IWSLT 2015}.

\bibitem[{Pan and Yang(2010)}]{Pan:IEEE2010}
Sinno~Jialin Pan and Qiang Yang. 2010.
\newblock A survey on transfer learning.
\newblock \emph{{IEEE} Trans. Knowl. Data Eng.}, 22(10).

\bibitem[{Papineni et~al.(2002)Papineni, Roukos, Ward, and
  Zhu}]{Papineni:ACL2002}
Kishore Papineni, Salim Roukos, Todd Ward, and Wei-Jing Zhu. 2002.
\newblock Bleu: a method for automatic evaluation of machine translation.
\newblock In \emph{Proc. of ACL 2002}.

\bibitem[{Pryzant et~al.(2017)Pryzant, Britz, and Le}]{Pryzant:WMT2017}
Reid Pryzant, Denny Britz, and Q~Le. 2017.
\newblock Effective domain mixing for neural machine translation.
\newblock In \emph{Proc. of WMT 2017}.

\bibitem[{Sajjad et~al.(2017)Sajjad, Durrani, Dalvi, Belinkov, and
  Vogel}]{Sajjad:arXiv2017}
Hassan Sajjad, Nadir Durrani, Fahim Dalvi, Yonatan Belinkov, and Stephan Vogel.
  2017.
\newblock Neural machine translation training in a multi-domain scenario.
\newblock \emph{CoRR abs/1708.08712}.

\bibitem[{Sennrich et~al.(2016)Sennrich, Haddow, and Birch}]{Sennrich:ACL2016}
Rico Sennrich, Barry Haddow, and Alexandra Birch. 2016.
\newblock Neural machine translation of rare words with subword units.
\newblock In \emph{Proc. of ACL 2016}.

\bibitem[{Servan et~al.(2016)Servan, Crego, and Senellart}]{Servan:Arxiv2016}
Christophe Servan, Josep Crego, and Jean Senellart. 2016.
\newblock Domain specialization: a post-training domain adaptation for neural
  machine translation.
\newblock \emph{CoRR abs/1612.06141.}

\bibitem[{Song et~al.(2019)Song, Gildea, Zhang, Wang, and Su}]{Song:TACL2019}
Linfeng Song, Daniel Gildea, Yue Zhang, Zhiguo Wang, and Jinsong Su. 2019.
\newblock Semantic neural machine translation using {AMR}.
\newblock \emph{TACL 2019}, 7.

\bibitem[{Su et~al.(2018)Su, Zeng, Xiong, Liu, Wang, and Xie}]{Su:TASLP2018}
Jinsong Su, Jiali Zeng, Deyi Xiong, Yang Liu, Mingxuan Wang, and Jun Xie. 2018.
\newblock A hierarchy-to-sequence attentional neural machine translation model.
\newblock \emph{{IEEE/ACM} TALSP 2018}, 26(3).

\bibitem[{Tan et~al.(2019)Tan, Ren, He, Qin, Zhao, and Liu}]{Tan:2019}
Xu~Tan, Yi~Ren, Di~He, Tao Qin, Zhou Zhao, and Tie{-}Yan Liu. 2019.
\newblock Multilingual neural machine translation with knowledge distillation.
\newblock In \emph{Proc. of ICLR 2019}.

\bibitem[{Tars and Fishel(2018)}]{Tars:arXiv2018}
Sander Tars and Mark Fishel. 2018.
\newblock Multi-domain neural machine translation.
\newblock \emph{CoRR abs/1805.02282.}

\bibitem[{Vaswani et~al.(2017)Vaswani, Shazeer, Parmar, Uszkoreit, Jones,
  Gomez, Kaiser, and Polosukhin}]{Vaswani:NIPS2017}
Ashish Vaswani, Noam Shazeer, Niki Parmar, Jakob Uszkoreit, Llion Jones,
  Aidan~N. Gomez, Lukasz Kaiser, and Illia Polosukhin. 2017.
\newblock Attention is all you need.
\newblock In \emph{Proc. of NIPS 2017}.

\bibitem[{Vilar(2018)}]{David:NAACL2018}
David Vilar. 2018.
\newblock Learning hidden unit contribution for adapting neural machine
  translation models.
\newblock In \emph{Proc. of NAACL 2018}, pages 500--505.

\bibitem[{Wang et~al.(2017{\natexlab{a}})Wang, Finch, Utiyama, and
  Sumita}]{Wang:ACL2017}
Rui Wang, Andrew Finch, Masao Utiyama, and Eiichiro Sumita. 2017{\natexlab{a}}.
\newblock Sentence embedding for neural machine translation domain adaptation.
\newblock In \emph{Proc. of ACL 2017}.

\bibitem[{Wang et~al.(2017{\natexlab{b}})Wang, Utiyama, Liu, Chen, and
  Sumita}]{Wang:EMNLP2017}
Rui Wang, Masao Utiyama, Lemao Liu, Kehai Chen, and Eiichiro Sumita.
  2017{\natexlab{b}}.
\newblock Instance weighting for neural machine translation domain adaptation.
\newblock In \emph{Proc. of EMNLP 2017}.

\bibitem[{Wang et~al.(2018)Wang, Xia, Zhao, Bian, Qin, Liu, and
  Liu}]{Wang:AAAI2018}
Yijun Wang, Yingce Xia, Li~Zhao, Jiang Bian, Tao Qin, Guiquan Liu, and
  Tie{-}Yan Liu. 2018.
\newblock Dual transfer learning for neural machine translation with marginal
  distribution regularization.
\newblock In \emph{Proc. of AAAI 2018}.

\bibitem[{Wu et~al.(2016)Wu, Schuster, Chen, Le, Norouzi, Macherey, Krikun,
  Cao, Gao, Macherey, Klingner, Shah, Johnson, Liu, Kaiser, Gouws, Kato, Kudo,
  Kazawa, Stevens, Kurian, Patil, Wang, Young, Smith, Riesa, Rudnick, Vinyals,
  Corrado, Hughes, and Dean}]{Wu:Arxiv2016}
Yonghui Wu, Mike Schuster, Zhifeng Chen, Quoc~V. Le, Mohammad Norouzi, Wolfgang
  Macherey, Maxim Krikun, Yuan Cao, Qin Gao, Klaus Macherey, Jeff Klingner,
  Apurva Shah, Melvin Johnson, Xiaobing Liu, Lukasz Kaiser, Stephan Gouws,
  Yoshikiyo Kato, Taku Kudo, Hideto Kazawa, Keith Stevens, George Kurian,
  Nishant Patil, Wei Wang, Cliff Young, Jason Smith, Jason Riesa, Alex Rudnick,
  Oriol Vinyals, Greg Corrado, Macduff Hughes, and Jeffrey Dean. 2016.
\newblock Google's neural machine translation system: Bridging the gap between
  human and machine translation.
\newblock \emph{CoRR abs/1609.08144}.

\bibitem[{Zeng et~al.(2018)Zeng, Su, Wen, Liu, Xie, Yin, and
  Zhao}]{Zeng:EMNLP2018}
Jiali Zeng, Jinsong Su, Huating Wen, Yang Liu, Jun Xie, Yongjing Yin, and
  Jianqiang Zhao. 2018.
\newblock Multi-domain neural machine translation with word-level domain
  context discrimination.
\newblock In \emph{Proc. of EMNLP 2018}.

\bibitem[{Zhang et~al.(2019{\natexlab{a}})Zhang, Shapiro, Kumar, McNamee,
  Carpuat, and Duh}]{zhang:NAACL2019}
Xuan Zhang, Pamela Shapiro, Gaurav Kumar, Paul McNamee, Marine Carpuat, and
  Kevin Duh. 2019{\natexlab{a}}.
\newblock Curriculum learning for domain adaptation in neural machine
  translation.
\newblock In \emph{Proc. of NAACL 2019}.

\bibitem[{Zhang et~al.(2019{\natexlab{b}})Zhang, Wu, Liu, Li, Zhou, and
  Chen}]{Zhang:AAAI2019}
Zhirui Zhang, Shuangzhi Wu, Shujie Liu, Mu~Li, Ming Zhou, and Enhong Chen.
  2019{\natexlab{b}}.
\newblock Regularizing neural machine translation by target-bidirectional
  agreement.
\newblock In \emph{Proc. of AAAI 2019}.

\bibitem[{Zoph et~al.(2016)Zoph, Yuret, May, and Knight}]{Zoph:EMNLP2016}
Barret Zoph, Deniz Yuret, Jonathan May, and Kevin Knight. 2016.
\newblock Transfer learning for low-resource neural machine translation.
\newblock \emph{Proc. of EMNLP 2016}.

\end{thebibliography}
\bibliographystyle{acl_natbib}
	
\appendix

\end{document}